\documentclass[aip,preprint]{revtex4-1}%linenumbers
\usepackage{graphicx}% Include figure files
\usepackage[section]{placeins}
 \usepackage{float}
 \usepackage{amsmath}

\begin{document}

% Use the \preprint command to place your local institutional report number 
% on the title page in preprint mode.
% Multiple \preprint commands are allowed.
%\preprint{}

\title{Noise robust neural network architecture}
\date{\today}

\author{Yunuo Xiong}
\email{xiongyunuo@hbpu.edu.cn}
\affiliation{Center for Fundamental Physics and School of Mathematics and Physics, Hubei Polytechnic University, Huangshi 435003, China}

\author{Hongwei Xiong}
%\email{xionghongwei@hbpu.edu.cn}

\affiliation{Center for Fundamental Physics and School of Mathematics and Physics, Hubei Polytechnic University, Huangshi 435003, China}

\date{\today}

\begin{abstract}
In which we propose neural network architecture (dune neural network) for recognizing general noisy image without adding any artificial noise in the training data. By representing each free parameter of the network as an uncertainty interval, and applying a linear transformation to each input element, we show that the resulting architecture achieves decent noise robustness when faced with input data with white noise. We apply simple dune neural networks for MNIST dataset and demonstrate that even for very noisy input images which are hard for human to recognize, our approach achieved better test set accuracy than human without dataset augmentation. 
We also find that our method is robust for many other examples with various background patterns added.

\end{abstract}

\maketitle

\section{Introduction}
The problem of devising neural network architectures that can recognize noisy image without artifically adding any noise in the training process has been long standing in the field of deep learning \cite{Jala,Koz,Dodge,Ziy,Baus,Zhou,Fag,syf,Nazare,Geirhos,Dodge1,Karki,Tang,Konar}. For example, it has been observed that state-of-the-art neural networks for ImageNet failed completely when we add small and special noises to the test images (the so-called adversarial examples \cite{Sze,Good,Ngu,Kurakin,kura2,Xie2,Sha,Xie,Pap,Atha,Metz,Zhou1,Guo}), and no efficient solutions have been found to tackle this issue and the presence of general noises. Therefore, noisy image recognition remains an active area of research \cite{Jala,Koz,Dodge,Ziy,Baus,Zhou,Fag,syf,Nazare,Geirhos,Dodge1,Karki,Tang,Konar,Sze,Good,Ngu,Kurakin,kura2,Xie2,Sha,Xie,Pap,Atha,Metz,Zhou1,Guo} and in this work, we propose a different neural network architecture which takes the effect of noises explicitly into account.

%For example, it has been observed that state-of-the-art neural networks for ImageNet failed completely when we add small and special noises to the test images, and no efficient and general solutions have been found to tackle this issue. \textbf{It is declared by Nielsn  that “it is premature to say that we're near solving the problem of image recognition.”} Therefore, \textbf{noisy image recognition or the so called adversarial images \cite{Kurakin} remain} an active area of research, and in this work, we propose a different neural network architecture which takes the effect of noises explicitly into account. 

In the dune neural network we present here, each free parameter of the model is represented as an uncertainty interval. At each training iteration, we sample from dune neural network using Monte Carlo method and propagate through the sampled network in the usual manner, then uncertainty intervals in the original dune network are dynamically updated in response to the loss derivatives. The advantages of using dune neural networks are twofold: first, the uncertainty interval representation of free parameters tends to make the model easier to optimize, due to how finite width trajectories make the exploration of loss function landscape easier, as contrast to other “probabilistic” models such as Bayesian neural network \cite{Jospin}, where model optimization proved to be much harder than optimization for standard architectures. Second, dune neural networks naturally encode robustness with respect to parameter perturbation, namely, all parameters in the given region defined by the uncertainty intervals should yield networks with similar performance. This property helps ensure that the training process is more stable as well.
\par
Inspired by the notion of uncertainty interval for dune neural networks, we also propose applying a linear transformation to each element of the noisy input data. This transformation does not eliminate the noise added to input data in any way, but maps the input uncertainty interval into a new uncertainty interval that achieves excellent noise robustness when fed into dune neural network. When we combine the input transformation with dune neural networks, we show that without dataset augmentation, the network realized high accuracy when evaluated on the noisy test set. Specifically, without dataset augmentation, in the examples we consider here, the noisy test set accuracy decreases roughly exponentially with increasing noise to signal ratio for traditional method with simple neural network; whereas for our approach here, the noisy test set accuracy only decreases approximately linearly. Therefore, we are able to achieve reasonable accuracy for very noisy images that are otherwise hard for a human to recognize, by training on the original training set only without the use of dataset augmentation. The work we present here may have potential value for the field of noisy image recognition and deep learning noise problem in general, and in particular may have a wide range of practical applications where we do not know or can not train the noises/disturbing background patterns in advance, e.g., in medical image recognition \cite{Shen} and astronomical image classification \cite{Kre,Sen}. 
\par
In the following, we give detailed account of dune neural network and its associated transformation of noisy input data, then we perform tests on MNIST dataset for simple neural network architectures to demonstrate the power of our approach to combat noisy data and disturbing background pattern.

\section{Method}

\subsection{Architecture}

Dune neural network converts each free parameter in an ordinary neural network into an uncertainty interval represented by two separate parameters. Let $\mathbf{\theta}$ denotes the collection of free parameters from the original network, then the corresponding dune neural network has parameters $\mathbf{\Theta}$ where $[\Theta_{i1},\Theta_{i2}]$ is the uncertainty interval corresponding to the $i$th free parameter $\theta_i$. So if the original network has $n$ parameters, the associated dune neural network has $2n$ parameters. Whenever we need to perform any operations on the dune neural network, we start by treating the uncertainty intervals as uniform distributions and sampling to obtain a set of real values, then construct an ordinary neural network based on those values:
\begin{equation}
\tilde{\theta_i}\in [\Theta_{i1},\Theta_{i2}].
\end{equation}
This is called the instantiation step. 

To perform forward propagation on dune neural network, we simply carry out the propagation on the sampled, ordinary neural network. For example, the loss function can be written as
\begin{equation}
L_{\mathbf{\Theta}}(\tilde{\theta_1},...,\tilde{\theta_n}).
\end{equation}
To update the uncertainty intervals of a dune neural network dynamically based on the loss function, we first calculate the gradient with respect to the sampled variables in the usual manner with backward propagation:
\begin{equation}
\mathbf{g}=\nabla_\mathbf{\tilde{\theta}}L_{\mathbf{\Theta}}(\tilde{\theta_1},...,\tilde{\theta_n}).
\end{equation}
Then we may use any optimization algorithm (stochastic gradient descent, momentum, Adam, and so forth) we want to calculate an update on $\mathbf{\tilde{\theta}}$ values based on the gradient and other relevant hyperparameters. Let us denote that step as
\begin{equation}
{\tilde{\bf{\theta}}^{(new)}}=optimization\_algorithm({\tilde{\bf{\theta}}^{(old)}},\mathbf{g},...).
\end{equation}

Then the parameters in the corresponding dune neural network are updated as
\begin{equation}
\nonumber
\Theta_{i1}^{(new)}=\Theta_{i1}^{(old)}+(1-p)(\tilde{\theta_i}^{(new)}-\tilde{\theta_i}^{(old)}),
\end{equation}
\begin{equation}
\Theta_{i2}^{(new)}=\Theta_{i2}^{(old)}+p(\tilde{\theta_i}^{(new)}-\tilde{\theta_i}^{(old)}),
\end{equation}
where $p$ is defined as
\begin{equation}
p=\frac{\tilde{\theta_i}^{(old)}-\Theta_{i1}^{(old)}}{\Theta_{i2}^{(old)}-\Theta_{i1}^{(old)}}.
\end{equation}
In case $\Theta_{i2}^{(old)}=\Theta_{i1}^{(old)}$ and $p$ is undefined, we use the following rule to update $\mathbf{\Theta}$
\begin{equation}
\nonumber
\Theta_{i1}^{(new)}=\Theta_{i1}^{(old)}+(\tilde{\theta_i}^{(new)}-\tilde{\theta_i}^{(old)}),
\end{equation}
\begin{equation}
\Theta_{i2}^{(new)}=\Theta_{i2}^{(old)}+(\tilde{\theta_i}^{(new)}-\tilde{\theta_i}^{(old)}).
\end{equation}
This ensures that when there are no uncertainty intervals (i.e. $\Theta_{i2}^{(old)}=\Theta_{i1}^{(old)}$), the usual parameter update rule for deterministic neural networks is recovered. 

The dynamics of uncertainty intervals in dune neural network resembles the movement of a dune in high dimensional space, hence the name. At each training iteration of dune neural network, we resample the parameter values from each uncertainty interval for better statistics. During testing process, we make a single sample from the dune neural network and treat the output of that sampled ordinary network as the final output. Usually, there is little gain from making multiple samples for a single iteration of dune neural network which decreases the efficiency. In its current form, dune neural networks have the same efficiency as traditional deterministic neural networks, and in deeper networks, dune neural networks have a good chance to converge to a solution faster than other networks under the same circumstances, due to the introduction of uncertainty intervals making it easier to explore the cost function landscape.
\par
To initialize a dune neural network, we first obtain a collection of parameters $\mathbf{\theta}$ for an ordinary network from any initialization scheme. Then we specify a prior uncertainty width $d$ and $\mathbf{\Theta}$ are set as
\begin{equation}
\Theta_{i1}=\theta_i-d,\ \Theta_{i2}=\theta_i+d.
\end{equation}
Also, to prevent any uncertainty interval from growing too large during training, we can add a regularization term for the width of each uncertainty interval to the loss function, as follows:
\begin{equation}
L^{(reg)}=L+\sum_{i=1}^n \beta (\Theta_{i2}-\Theta_{i1})^2,
\end{equation}
where $\beta$ is a hyperparameter similar to the one from weight decay. And to prevent the uncertainty interval from growing too small, which would recover a deterministic network, we can impose a hard threshold for the smallest allowed uncertainty interval width, by specifying another hyperparameter $w_{min}$. If after an update is made, an uncertainty interval has a width less than $w_{min}$ (i.e. $\Theta_{i2}-\Theta_{i1}<w_{min}$), then we readjust that uncertainty interval as
\begin{equation}
\nonumber
\Theta_{i1}=\frac{\Theta_{i1}+\Theta_{i2}}{2}-\frac{w_{min}}{2},
\end{equation}
\begin{equation}
\Theta_{i2}=\frac{\Theta_{i1}+\Theta_{i2}}{2}+\frac{w_{min}}{2}.
\end{equation}

\subsection{Data representation}
In image recognition, we normalize the input data so that each pixel is represented as a single real value in the interval $x\in[0,1]$ for grayscale images. For color images each pixel would be represented as three numbers each in the interval $[0,1]$. In this work, we first consider white noise as applied to the input images, by using this formula to add noise to all pixels in the image:
\begin{equation}
\label{noise}
x^{noisy}=\frac{x^{old}}{n_s+1}+\frac{n_s}{n_s+1}U(0,1),
\end{equation}
where $U(0,1)$ denotes a random number from the uniform distribution $[0,1]$. 

It is easy to see that the noise-to-signal ratio in the above formula is $n_s$, the larger $n_s$ the more noise is applied to the image. Regardless of whether there are noises present in the image, each pixel is always restricted to lie in the interval $[0,1]$. Based on the idea of dune neural network, we may treat the range of pixel values $[0,1]$ as an uncertainty interval too. In particular, we propose to apply the same linear transformation to each pixel value, before feeding those values as input to the dune neural network, with the following formula:
\begin{equation}
\label{magics}
\tilde{x}^{noisy}=(1+2h_s)x^{noisy}-h_s.
\end{equation}
It is straightforward to see that $\tilde{x}^{noisy}$ now lies in the interval $[-h_s,1+h_s]$. 

It is also worthy to note that the above transformation does not eliminate or weaken the noise in the noisy input data in any way, since the transformation is applied after the noise adding procedure given by Eq. (\ref{noise}). The $h_s$ in the above equation is to be treated as a hyperparameter specified by the user. We will see later that after applying the transformation to the input data, the capability of dune neural network to recognize noisy images is dramatically improved. We call the input data representation as defined by Eq. (\ref{magics}) magic shift (abbreviated Magics).
\par
In the following, we apply the approach introduced in this section to MNIST dataset with noisy test images. We show that even without dataset augmentation, our architecture and data representation achieved excellent noise robustness compared with previous methods \cite{Jala,Nazare,Tang}.

\section{Results}
We test our approach on MNIST dataset for simple dune neural networks. Specifically, we use dune network counterpart of simple fully connected network for image classification to verify the principle of our method, with ReLU activation and softmax classifier. We do not use dataset augmentation so the training set always remains the same, and when we need to add noise to the testing set, we add white noise using Eq. (\ref{noise}) ($n_s=0$ corresponds to the case of no noise).
\par

\begin{figure}[htbp]
\begin{center}
\includegraphics[width=0.75\textwidth]{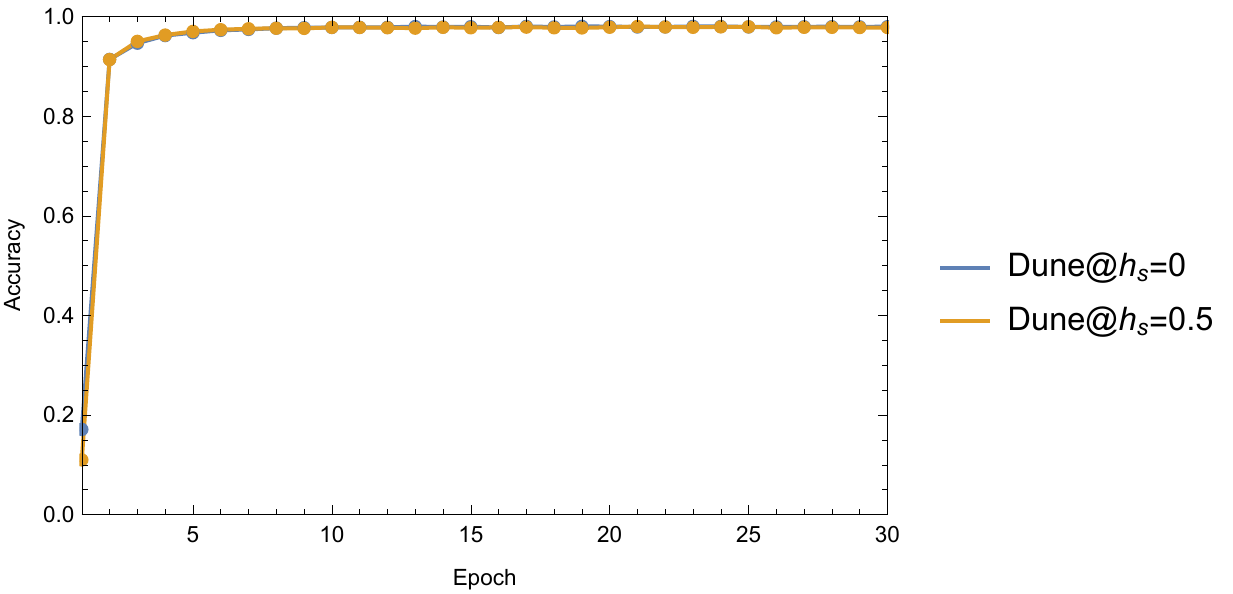} 
\caption{\label{Verify} Test set (without noise) accuracy as a function of epochs for two layer fully connected dune neural network of hidden units 150 and 10. We used Adam optimization algorithm and the learning rate is 0.001, $\beta=0.1$ and $w_{min}=0.15$. In this simple setting, both the cases with Magics ($h_s=0.5$) and without Magics ($h_s=0$) converge toward 100\% accuracy. }
\end{center}
\end{figure}

To begin with, we check that the combination of dune neural networks and Magics is working correctly, by monitoring the accuracy on the original testing set (without noise). The result is shown in Fig. \ref{Verify}; as expected, the accuracy converges toward 100\% at a stable pace.
\par

\begin{figure}[htbp]
\begin{center}
\includegraphics[width=0.75\textwidth]{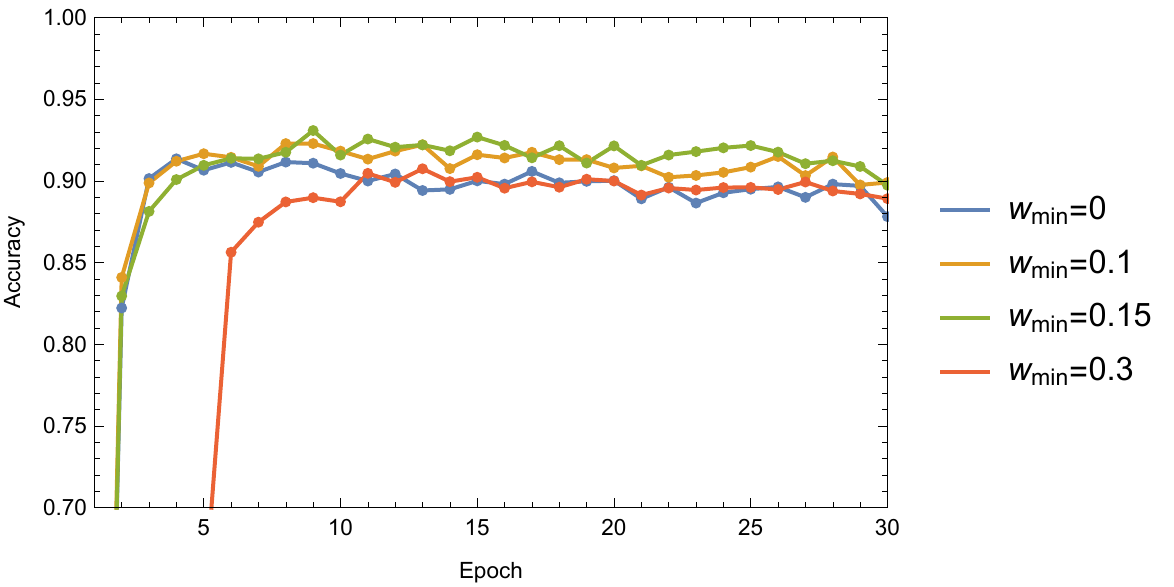} 
\caption{\label{Wmin} Noisy test set accuracy ($n_s=1.5$, no dataset augmentation) as a function of epochs for two layer fully connected dune neural network of hidden units 150 and 10, for different $w_{min}$ values. The common hyperparameters are: learning rate 0.001, $\beta=0.1$, $h_s=1.8$. We can see that using different $w_{min}$ values can have a slight effect on the test set accuracy, with $w_{min}=0.15$ corresponding to the optimal choice. In practice one may need to experiment with different $w_{min}$ to find the best choice. }
\end{center}
\end{figure}

Then we turn our attentions to the case of noisy test set. In Fig. \ref{Wmin}, we tested dune neural network with Magics on noisy test set with $n_s=1.5$, without dataset augmentation. We give results for different $w_{min}$ values, we can see that $w_{min}=0.15$ gives optimal result.

\begin{figure}[htbp]
\begin{center}
\includegraphics[width=0.75\textwidth]{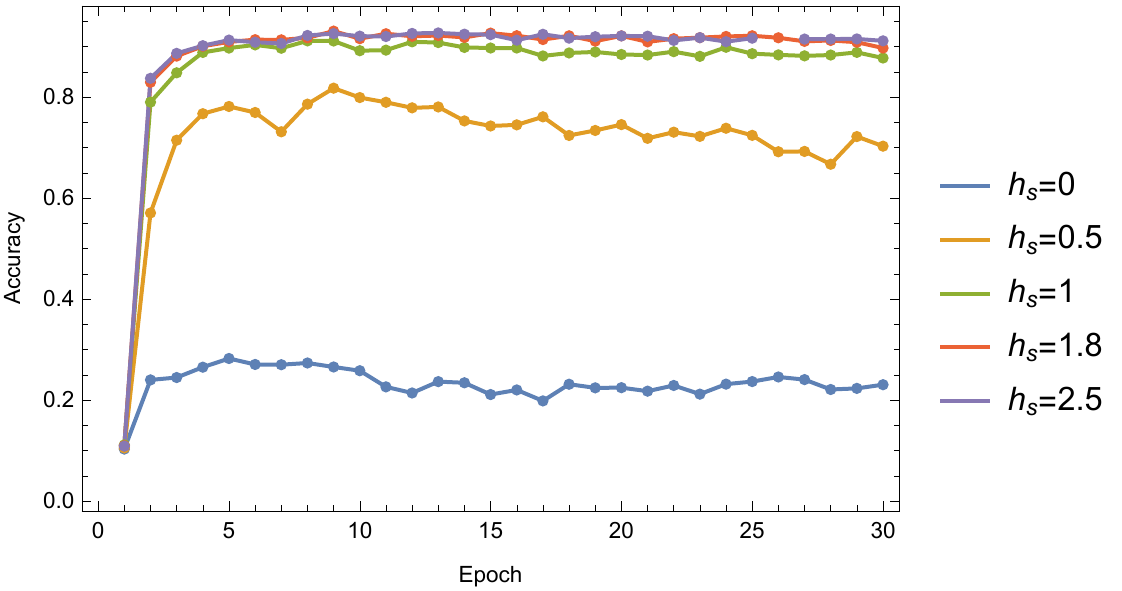} 
\caption{\label{Shift} Noisy test set accuracy ($n_s=1.5$, no dataset augmentation) as a function of epochs for two layer fully connected dune neural network of hidden units 150 and 10, for different $h_s$ values. The common hyperparameters are: learning rate 0.001, $\beta=0.1$, $w_{min}=0.15$. $h_s=0$ corresponds to the case without Magics and the network performs poorly. By using positive $h_s$ values, we see that there is a dramatical improvement in the noisy test set accuracy. With large enough $h_s$, the accuracy converges to about 91\%. }
\end{center}
\end{figure}

\par
Next, we investigate the effects of Magics on the noisy test set accuracy, without dataset augmentation. As we mentioned before, the combination of dune neural network and Magics can improve the ability to recognize noisy images dramatically, so we try different Magics hyperparameter $h_s$ ($h_s=0$ is the special case with no Magics). The result is shown in Fig. \ref{Shift}, we can clearly see the improvement in noisy test set accuracy by using positive $h_s$ values. As $h_s$ increases the accuracy begins to converge.

\begin{figure}[htbp]
\begin{center}
\includegraphics[width=0.75\textwidth]{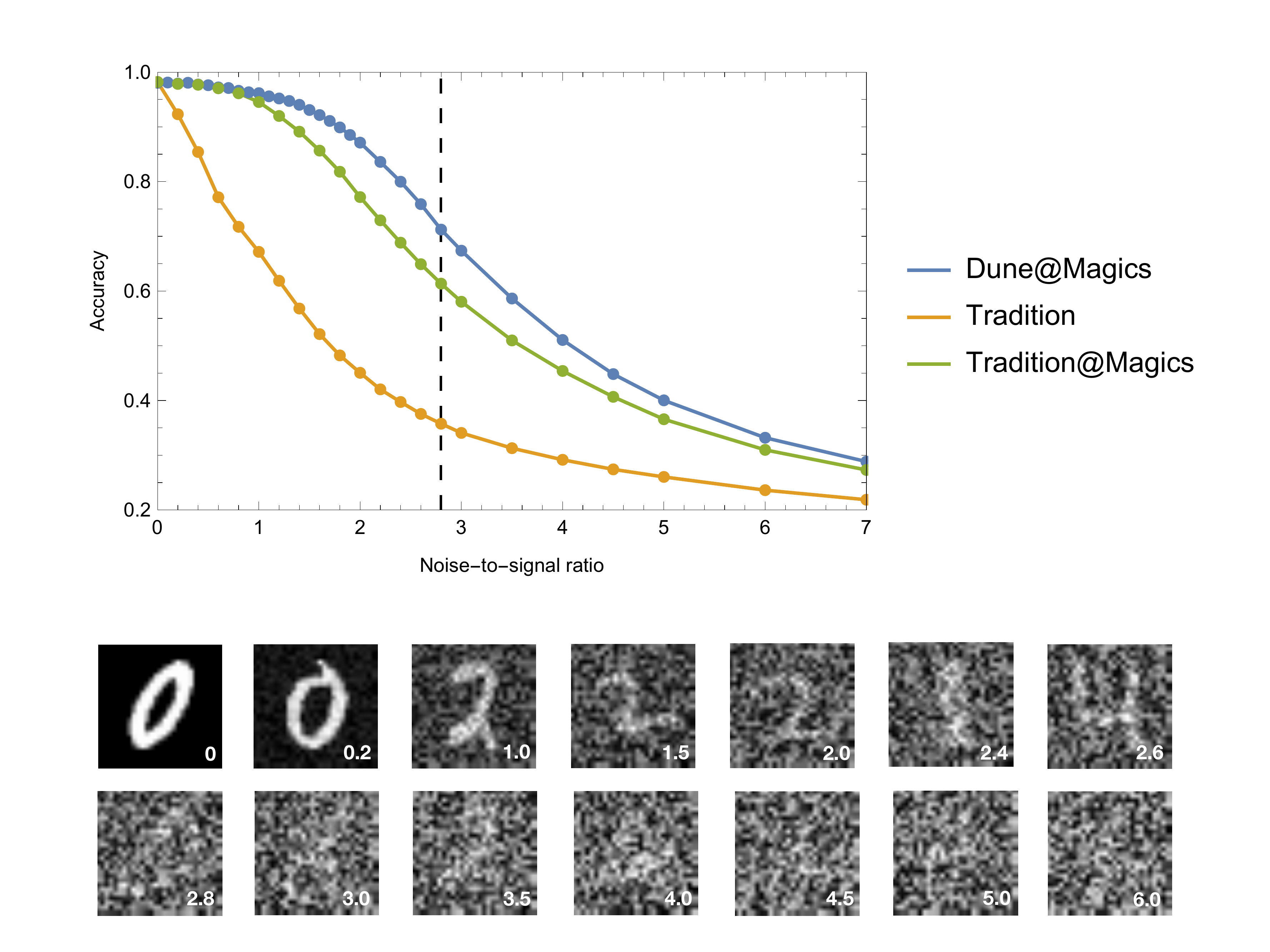} 
\caption{\label{MnistNoise} Noisy test set accuracy as a function of noise-to-signal ratio $n_s$ (no dataset augmentation) for two layer fully connected dune neural network of hidden units 150 and 10, for different methods. The hyperparameters are: learning rate 0.001, $\beta=0.1$, $w_{min}=0.15$. For $n_s\geq1$, we set $h_s=1.8$, otherwise we set $h_s=2n_s$. Traditional method shown by yellow line (without dune neural network and without Magics) has a poor performance: with increasing noise-to-signal ratio the test set accuracy decreases exponentially. The combination of dune neural network and Magics (shown by blue line) achieves the best performance, while the combination of ordinary neural network and Magics (shown by green line) doing slightly worse. The dashed line in the figure corresponds to the threshold of human's ability to recognize noisy images, we show a selection of images from MNIST for different $n_s$ in the bottom of the figure. }
\end{center}
\end{figure}

\begin{figure}[htbp]
\begin{center}
\includegraphics[width=0.75\textwidth]{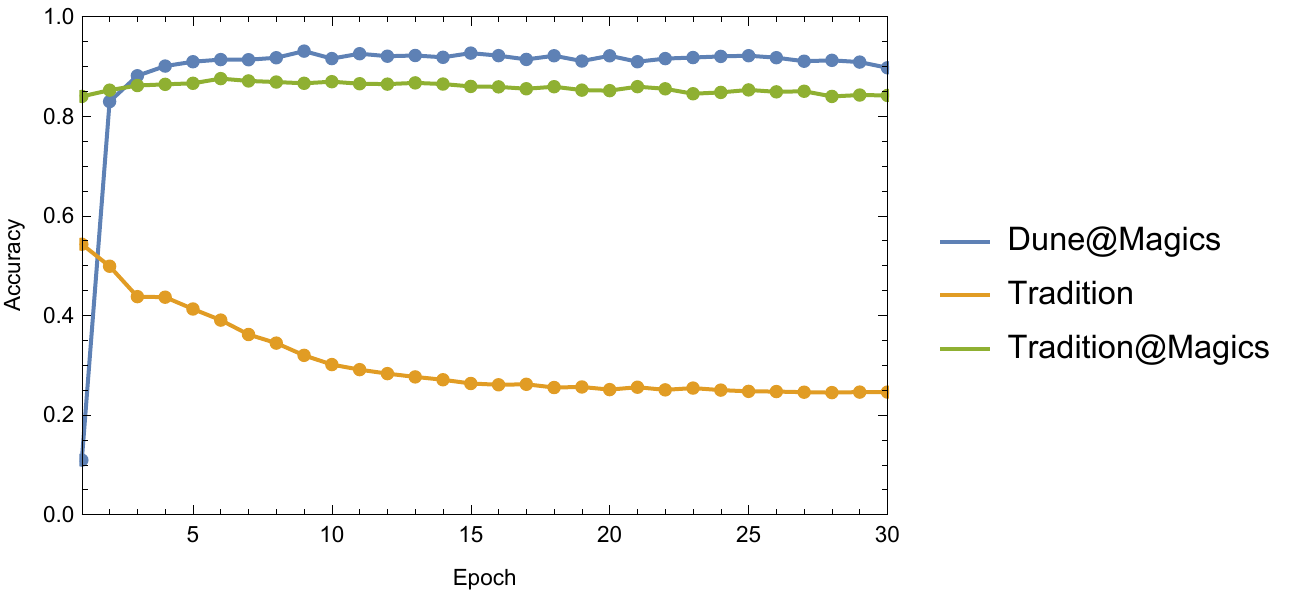} 
\caption{\label{Convergence} Noisy test set accuracy ($n_s=1.5$, no dataset augmentation) as a function of epochs for two layer fully connected dune neural network of hidden units 150 and 10, for different methods. The hyperparameters are: learning rate 0.001, $\beta=0.1$, $w_{min}=0.15$. For the two methods with Magics, we set $h_s=1.8$. }
\end{center}
\end{figure}

\par
We note that the traditional method with simple neural network can be combined with Magics as well. To compare the noise robustness of these different approaches, we plot the noisy training set accuracy versus noise-to-signal ratio for three methods, as shown in Fig. \ref{MnistNoise}. As shown by the yellow line based on the traditional method, the test set accuracy decreases exponentially with increasing noise-to-signal ratio. This exponential decay is also observed in many other works \cite{Dodge,Ziy} with more complex neural network, e.g. for Gaussian blur and Gaussian noise in Ref. \cite{Dodge}. One of the most challenging problem in noisy image recognition is to eliminate this exponentially decaying behavior, which is successfully realized in Fig. \ref{MnistNoise}, as shown by blue line based on the combination of dune neural network and Magics, and the green line based on the combination of the traditional method and Magics.

We also pick a specific case and plot the accuracy as a function of epochs for these three different methods, as shown in Fig. \ref{Convergence}. We can see that the accuracy of traditional method decreases quickly with epochs due to overfitting, whereas the two methods with Magics both achieved decent noise robustness and did not overfit over the training course of 30 epochs.

\begin{figure}[htbp]
\begin{center}
\includegraphics[width=0.75\textwidth]{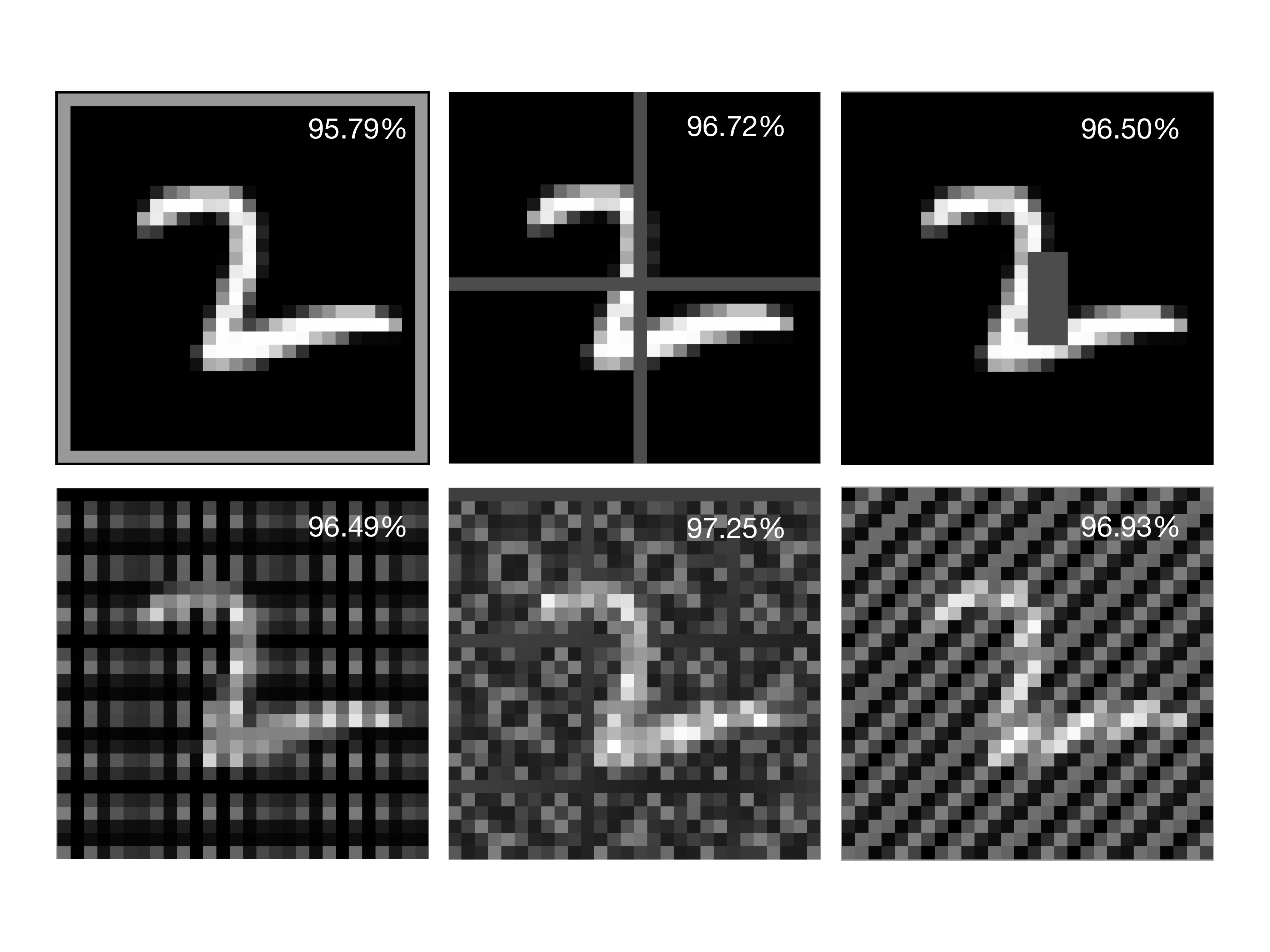} 
\caption{\label{Mask} The same image with different types of noises applied to it, the upper left corner is the original image with added border. At the top right corner of each image, we show the noisy test set accuracy (without dataset augmentation), using the combination of dune neural network and Magics. A decent noise robustness is achieved in each case, showing that our method here can be used to deal with a variety of different noises and disturbing background patterns. }
\end{center}
\end{figure}

\par
We emphasize that in our method there is no artificial noise added in the training dataset for recognizing test images with added white noise. Hence, it is expected that our method can be applied to other types of noise or background pattern.  To support this conjecture, we also performed experiments on different types of noises and found that the combination of dune neural network and Magics achieved decent noise robustness for all the different noises we considered. A selection of images with various disturbing background patterns along with the test set accuracy are shown in Fig. \ref{Mask}. The improvement on the test set accuracy of noisy image recognition is significant, compared with previous results with more complex methods \cite{Jala,Nazare,Tang}.

\section{Conclusion}
As a summary, in this work we propose a noise robust neural network architecture, based on the combination of dune neural network and Magics as applied to the input data. We tested on MNIST for different noises and noise strength to demonstrate the improvement of the current method over traditional method. In particular, we achieved decent noisy test set accuracy even without dataset augmentation, showing that our method is indeed noise robust for noisy image recognition.

\par
We believe the method present here has a broad applicability when faced with general noise problem in the field of deep learning. For example, all existing neural network architectures can be promoted to their dune neural network version without much extra difficulty, so our approach can naturally be applied to tasks other than image classification in the future. Moreover, a possible future direction is to gain a deeper understanding of Magics, by applying it to other kinds of noisy input data to see if we can achieve noise robustness in other applications.
The purpose of the present work is not to consider adversarial examples \cite{Sze,Good,Ngu,Kurakin,kura2,Xie2,Sha,Xie,Pap,Atha,Metz,Zhou1,Guo}, which may deserve further studies based on our method.

\textbf{Acknowledgments}  The authors would like to acknowledge support from Hubei Polytechnic University.

\textbf{Author Contributions} YN  designed the study, proposed the idea, implemented the methods and wrote the manuscript. HW performed some numerical experiments and revised the manuscript.

\textbf{Funding} This work is partly supported by the National Natural Science Foundation of China under grant numbers 11175246, and 11334001. 

\textbf{Declaration of competing interest}

The authors declare that they have no known competing financial interests or personal relationships that could have appeared to influence the work reported in this paper.

\textbf{Availability of data and material}
The data that support the findings of this study are available from the corresponding author upon reasonable request.

\textbf{Declarations}

\textbf{Confict of interest} The authors declare no conflicts of interest.

\textbf{Ethics approval} Not Applicable.

\textbf{Consent to participate} Not Applicable.

\textbf{Consent for publication} Not Applicable.

\textbf{Code availability} All code associated with this paper is publicly available from https://github.com/xiongyunuo/XNN


\begin{thebibliography}{10}

% Noisy image recognition
\bibitem{Jala} Jalalvand, A., Demuynck, K., De Neve, W.,  Martens, J. P. (2018). On the application of reservoir computing networks for noisy image recognition. Neurocomputing, 277, 237-248.

\bibitem{Koz} Koziarski, M.,  Cyganek, B. (2017). Image recognition with deep neural networks in presence of noise–dealing with and taking advantage of distortions. Integrated Computer-Aided Engineering, 24(4), 337-349.

\bibitem{Dodge} Dodge, S., Karam, L. (2016). Understanding how image quality affects deep neural networks. In 2016 eighth international conference on quality of multimedia experience (QoMEX) (pp. 1-6). IEEE.

\bibitem{Ziy} Ziyadinov, V., Tereshonok, M. (2022). Noise immunity and robustness study of image recognition using a convolutional neural network. Sensors, 22(3), 1241.

\bibitem{Baus} Basu, S., Karki, M., Ganguly, S., DiBiano, R., Mukhopadhyay, S., Gayaka, S., ...  Nemani, R. (2017). Learning sparse feature representations using probabilistic quadtrees and deep belief nets. Neural Processing Letters, 45, 855-867.

\bibitem{Zhou} Zhou, Y., Song, S., Cheung, N. M. (2017). On classification of distorted images with deep convolutional neural networks. In 2017 IEEE International Conference on Acoustics, Speech and Signal Processing (ICASSP) (pp. 1213-1217). IEEE.

\bibitem{Fag} Fagbohungbe, O., Qian, L. (2022). The Effect of Batch Normalization on Noise Resistant Property of Deep Learning Models. IEEE Access, 10, 127728-127741.

\bibitem{syf} Roy, S. S., Hossain, S. I., Akhand, M. A. H., Murase, K. (2018). A robust system for noisy image classification combining denoising autoencoder and convolutional neural network. International Journal of Advanced Computer Science and Applications, 9(1), 224-235.

\bibitem{Nazare} Nazaré, T. S., da Costa, G. B. P., Contato, W. A.,  Ponti, M. (2018). Deep convolutional neural networks and noisy images. In Progress in Pattern Recognition, Image Analysis, Computer Vision, and Applications: 22nd Iberoamerican Congress, CIARP 2017, Valparaíso, Chile, November 7–10, 2017, Proceedings 22 (pp. 416-424). Springer International Publishing.

\bibitem{Geirhos} Geirhos, R., Janssen, D. H., Schütt, H. H., Rauber, J., Bethge, M.,  Wichmann, F. A. (2017). Comparing deep neural networks against humans: object recognition when the signal gets weaker. arXiv preprint arXiv:1706.06969.

\bibitem{Dodge1} Dodge, S.,  Karam, L. (2017). A study and comparison of human and deep learning recognition performance under visual distortions. In 2017 26th international conference on computer communication and networks (ICCCN) (pp. 1-7). IEEE.

\bibitem{Karki} Karki, M., Liu, Q., DiBiano, R., Basu, S., Mukhopadhyay, S. (2018). Pixel-level reconstruction and classification for noisy handwritten bangla characters. In 2018 16th International Conference on Frontiers in Handwriting Recognition (ICFHR) (pp. 511-516). IEEE.

\bibitem{Tang} Tang, Y.,  Eliasmith, C. (2010). Deep networks for robust visual recognition. In Proceedings of the 27th International Conference on Machine Learning (ICML-10) (pp. 1055-1062).

\bibitem{Konar} Konar, D., Gelenbe, E., Bhandary, S., Sarma, A. D., Cangi, A. (2022). Random quantum neural networks (RQNN) for noisy image recognition. arXiv preprint arXiv:2203.01764.


%adversarial images

\bibitem{Sze} Szegedy, C., Zaremba, W., Sutskever, I., Bruna, J., Erhan, D., Goodfellow, I., Fergus, R. (2013). Intriguing properties of neural networks. arXiv preprint arXiv:1312.6199.

\bibitem{Good} Goodfellow, I. J., Shlens, J.,  Szegedy, C. (2014). Explaining and harnessing adversarial examples. arXiv preprint arXiv:1412.6572.

\bibitem{Ngu} Nguyen, A., Yosinski, J., Clune, J. (2015). Deep neural networks are easily fooled: High confidence predictions for unrecognizable images. In Proceedings of the IEEE conference on computer vision and pattern recognition (pp. 427-436).

\bibitem{Kurakin} Kurakin, A., Goodfellow, I. J., Bengio, S. (2018). Adversarial examples in the physical world. In Artificial intelligence safety and security (pp. 99-112). Chapman and Hall/CRC.

\bibitem{kura2} Kurakin, A., Goodfellow, I.,  Bengio, S. (2016). Adversarial machine learning at scale. arXiv preprint arXiv:1611.01236.

\bibitem{Xie2} Xie, C., Wu, Y., Maaten, L. V. D., Yuille, A. L., He, K. (2019). Feature denoising for improving adversarial robustness. In Proceedings of the IEEE/CVF conference on computer vision and pattern recognition (pp. 501-509).

\bibitem{Sha} Shafahi, A., Najibi, M., Ghiasi, M. A., Xu, Z., Dickerson, J., Studer, C., ...,  Goldstein, T. (2019). Adversarial training for free!. Advances in Neural Information Processing Systems, 32.

\bibitem{Xie} Xie, C., Tan, M., Gong, B., Wang, J., Yuille, A. L.,  Le, Q. V. (2020). Adversarial examples improve image recognition. In Proceedings of the IEEE/CVF Conference on Computer Vision and Pattern Recognition (pp. 819-828).

\bibitem{Pap} Papernot, N., McDaniel, P., Jha, S., Fredrikson, M., Celik, Z. B.,  Swami, A. (2016). The limitations of deep learning in adversarial settings. In 2016 IEEE European symposium on security and privacy (EuroS P) (pp. 372-387). IEEE.

\bibitem{Atha} Athalye, A., Engstrom, L., Ilyas, A.,  Kwok, K. (2018). Synthesizing robust adversarial examples. In International conference on machine learning (pp. 284-293). PMLR.

\bibitem{Metz} Metzen, J. H., Genewein, T., Fischer, V., Bischoff, B. (2017). On detecting adversarial perturbations. arXiv preprint arXiv:1702.04267.

\bibitem{Zhou1} Zhou, Z.,  Firestone, C. (2019). Humans can decipher adversarial images. Nature communications, 10(1), 1334.

\bibitem{Guo} Guo, C., Rana, M., Cisse, M.,  Van Der Maaten, L. (2017). Countering adversarial images using input transformations. arXiv preprint arXiv:1711.00117.


% Book
%\bibitem{Nielsn} Nielsen, M. A. (2015). Neural networks and deep learning. San Francisco, CA, USA: Determination press.


%Bayesian neural networks
\bibitem{Jospin} Jospin, L. V., Laga, H., Boussaid, F., Buntine, W.,  Bennamoun, M. (2022). Hands-on Bayesian neural networks—A tutorial for deep learning users. IEEE Computational Intelligence Magazine, 17(2), 29-48.

%Medical image

\bibitem{Shen} Shen, D., Wu, G., Suk, H. I. (2017). Deep learning in medical image analysis. Annual review of biomedical engineering, 19, 221-248.

%Astronomical

\bibitem{Kre} Kremer, J., Stensbo-Smidt, K., Gieseke, F., Pedersen, K. S., Igel, C. (2017). Big universe, big data: machine learning and image analysis for astronomy. IEEE Intelligent Systems, 32(2), 16-22.

\bibitem{Sen} Sen, S., Agarwal, S., Chakraborty, P.,  Singh, K. P. (2022). Astronomical big data processing using machine learning: A comprehensive review. Experimental Astronomy, 53(1), 1-43.


\end{thebibliography}
\end{document}